\documentclass[10pt,twocolumn,letterpaper]{article}

\usepackage{cvpr}              
\usepackage{adjustbox}
\usepackage{multirow}
\usepackage{multicol}
\usepackage{float}
\usepackage{placeins}
\usepackage{wrapfig}

\definecolor{cvprblue}{rgb}{0.21,0.49,0.74}
\definecolor{ultramarine}{RGB}{0,32,96}
\usepackage[pagebackref,breaklinks,colorlinks,allcolors=cvprblue]{hyperref}
\usepackage[dvipsnames]{xcolor}
\def\paperID{12553} 
\def\confName{CVPR}
\def\confYear{2025}
\newcommand{\modelname}{\textsc{ICT}\xspace}

\newcommand{\squishend}{
\end{list} }
\newcommand{\squishlist}{
\begin{list}{$\bullet$}
{ \usecounter{Lcount}
\setlength{\itemsep}{0pt}
\setlength{\parsep}{0pt}
\setlength{\topsep}{0pt}
\setlength{\partopsep}{0pt}
\setlength{\leftmargin}{2em}
\setlength{\labelwidth}{1.5em}
\setlength{\labelsep}{0.5em} } }
\title{\modelname: Image-Object Cross-Level Trusted Intervention for Mitigating Object Hallucination in Large Vision-Language Models}

\author{\textbf{Junzhe Chen}$^{1,2*}$, \textbf{Tianshu Zhang}$^{1*}$, \textbf{Shiyu Huang}$^{3}$, \textbf{Yuwei Niu}$^{4}$, \\ \textbf{Linfeng Zhang}$^{5}$, 
\textbf{Lijie Wen}$^{1\dagger}$, \textbf{Xuming Hu}$^{2\dagger}$\\
  $^1$Tsinghua University, $^2$The Hong Kong University of Science and Technology (Guangzhou),\\
  $^3$Zhipu AI, $^4$Chongqing University, $^5$Shanghai Jiao Tong University.\\ \texttt{chenjz24@mails.tsinghua.edu.cn,}\\ \texttt{wenlj@tsinghua.edu.cn, xuminghu97@gmail.com,}
  }

\begin{document}
\maketitle
{
\let\thefootnote\relax\footnotetext{
$^*$ Equal Contribution.}
\let\thefootnote\relax\footnotetext{
$^\dagger$ Corresponding authors. }
}
\begin{abstract}
Despite the recent breakthroughs achieved by Large Vision Language Models (LVLMs) in understanding and responding to complex visual-textual contexts, their inherent hallucination tendencies limit their practical application in real-world scenarios that demand high levels of precision. Existing methods typically either fine-tune the LVLMs using additional data, which incurs extra costs in manual annotation and computational resources or perform comparisons at the decoding stage, which may eliminate useful language priors for reasoning while introducing inference time overhead. Therefore, we propose \modelname, a lightweight, training-free method that calculates an intervention direction to shift the model's focus towards different levels of visual information, enhancing its attention to high-level and fine-grained visual details.  During the forward pass stage, the intervention is applied to the attention heads that encode the overall image information and the fine-grained object details, effectively mitigating the phenomenon of overly language priors, and thereby alleviating hallucinations. Extensive experiments demonstrate that \modelname achieves strong performance with a small amount of data and generalizes well across different datasets and models. Our code will be public.
\end{abstract}    
\section{Introduction}
\begin{figure}
  \centering
  \includegraphics[width=\linewidth]{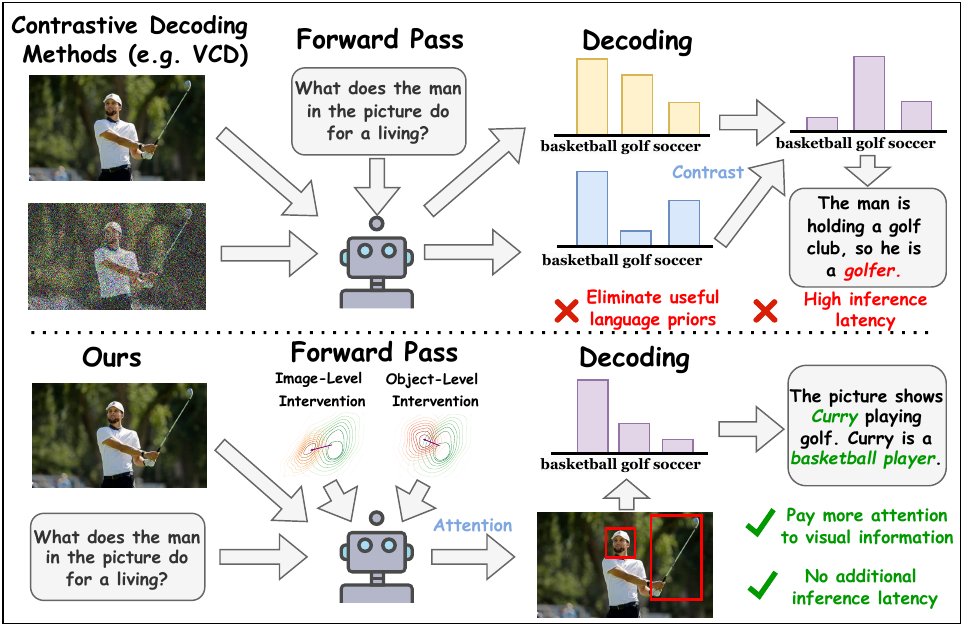}
  \vspace{-6mm}
  \caption{Comparison between Contrastive Decoding (top) and our proposed \modelname (bottom). In the top example, Contrastive Decoding indiscriminately removes both beneficial and detrimental language priors, leading to hallucinations. In contrast, our approach enhances the model's attention to visual details while preserving useful language priors, allowing it to correctly identify and describe objects in the image.}
  \label{figure:overview}
  \vspace{-8mm}
\end{figure}
The recent rapid advancements in Large Vision Language Models (LVLMs) have demonstrated significant potential to tackle complex real-world tasks~\cite{liu2023improvedllava, Bai2023,hong2024cogvlm2,liu2024kangaroo,hu2023look}. However, LVLMs frequently generate text outputs that are inconsistent with the visual input, such as incorrectly determining the presence of objects in an image or inaccurately describing the attributes of those objects~\cite{liu2024survey,sahoo2024comprehensive}. This issue is commonly referred to as the \textit{object hallucination phenomenon}~\cite{liu2024survey}, which significantly limits the applicability of LVLMs in high-stakes scenarios that demand precise accuracy, such as autonomous driving~\cite{yang2023llm4drive} and medical surgery~\cite{wang2024surgical,jin2024surgical}. Previous research suggested that the primary reasons for hallucinations in LVLMs are as follows: 1) Large language models (LLMs), such as Vicuna~\cite{chiang2023vicuna}, possess significantly stronger parameters and capabilities compared to visual encoders, such as CLIP~\cite{radford2021learning}. This results in an excessively strong language prior, causing the model to overly rely on language cues while neglecting visual input~\cite{liu2024survey, leng2024mitigating, han2022visual,lee2024vlind, gupta2022swapmix}. 2) Current visual decoders often struggle to accurately capture fine-grained visual semantics, resulting in errors in detailed object attributes in an image, such as color and quantity ~\cite{liu2024survey, chenhalc, an2024agla, cho2022fine}. 

Based on the two issues outlined above, previous efforts to mitigate the hallucination phenomenon in LVLMs can be categorized into three main approaches: 1) \textbf{Fine-tuning with additional data:} This approach involves introducing high-quality annotated data to better align the model’s behavior with human interpretations, effectively teaching the model to focus more on visual information ~\cite{zhao2023beyond,liu2023improvedllava,gunjal2024detecting,li2024vlfeedback}. However, this method not only requires costly human annotation but also involves updating model parameters, which demands substantial computational resources, thereby limiting its scalability. 2) \textbf{Perceptual enhancement}: This method incorporates additional information, such as depth maps and segmentation maps~\cite{jain2024vcoder,zhao2023beyond,lee2024multimodal}, as auxiliary inputs to assist the visual encoder in capturing more detailed visual features, thereby reducing hallucinations. However, it often requires manual selection of auxiliary features, which limits its generalizability across different tasks. 3) \textbf{Contrastive decoding:} This method alleviates hallucinations without requiring additional training. It induces hallucinations by applying transformations such as blurring, rotation, or cropping to the original visual input. During the decoding stage, tokens associated with these induced hallucinations are penalized, thus mitigating the influence of language priors ~\cite{leng2024mitigating, Zhu2024MultilingualContrastiveDecoding, Chen2024HALC, Chuang2024DoLa, Zhong2024, Kim2024, woo2024dontmissforesttrees, kim2024codecontrastingselfgenerateddescription, phan2024distillationcontrastivedecodingimproving}. However, methods such as VCD often indiscriminately eliminate all language priors, including those that may be beneficial. As shown in Figure \ref{figure:overview}, the original model recognizes Curry holding a golf club, while utilizing the language prior that identifies Curry as a basketball player. This language prior is valuable for achieving an accurate interpretation. However, through contrastive decoding, this useful prior is also removed, which can inadvertently result in hallucinations.

To address the challenges of mitigating hallucinations in LVLMs, we propose \textbf{I}mage-Object \textbf{C}ross-Level \textbf{T}rusted Intervention (\modelname), a training-free, plug-and-play method applicable during the forward pass. Unlike contrastive decoding, our approach does not eliminate language priors to reduce the model’s over-reliance on text modality. Instead, it intervenes during the forward pass to enhance the model’s focus on both comprehensive visual information and fine-grained object details. We explored the activation patterns of attention heads when the model produces correct versus hallucinated responses. This analysis allowed us to identify activation value deviations that can shift the model from being ``untrustworthy'' to ``trustworthy''. According to prior research \cite{voita2019analyzing, ferrando2024primerinnerworkingstransformerbased,chen2024selfieselfinterpretationlargelanguage}, which has shown that different heads in the multi-head attention mechanism encode information at varying levels of granularity, we train binary classifiers for each head to determine which heads encode overall visual information and which capture detailed visual features. During the forward pass, we adjust the activation values of these heads based on their identified levels, thereby enhancing the model’s attention to the relevant visual features and reducing the likelihood of hallucinations. As illustrated in Figure \ref{figure:overview}, after applying \modelname, the model is able to focus more on the details within the image, such as identifying the man as Curry, while simultaneously utilizing beneficial language priors (e.g., Curry is a basketball player) to infer and arrive at the correct answer. Since the intervention shift vectors are pre-computed, \modelname does not introduce additional delays during the forward pass.

Our experiments demonstrate that, for both LLaVA-v1.5~\cite{liu2023improvedllava} and Qwen-VL~\cite{Bai2023}, applying \modelname results in an average improvement of $6.27\%$ on POPE benchmark and $67.37$ points on MME benchmark. Furthermore, \modelname exhibits cross-dataset generalization and model-agnostic generalizability. Our contributions can be summarized as:

$\bullet$ We propose \textbf{I}mage-Object \textbf{C}ross-Level \textbf{T}rusted Intervention (\modelname), a novel, training-free, plug-and-play method that effectively reduces hallucinations in LVLMs by enhancing focus on both overall visual information and fine-grained object details during the forward pass, without eliminating beneficial language priors.

$\bullet$ Unlike existing contrastive decoding approaches, we introduce an intervention mechanism that operates during the forward pass rather than the decoding stage, allowing \modelname to be orthogonal and complementary to existing decoding strategies while not introducing any additional latency.

$\bullet$ Extensive experiments on LLaVA-v1.5 and Qwen-VL demonstrate that \modelname significantly improves performance on both the POPE and MME benchmarks while maintaining cross-dataset and model-agnostic generalizability.

\section{Related Work}

\subsection{Large Vision-Language Models}
Following the success of LLM \citep{Brown2020, Llama2023, gpt4, GLM2024, chiang2023vicuna}, researchers have been exploring the multimodal field. Taking advantage of the remarkable capabilities of LLMs, Large Vision-Language Models incorporate visual encoders and feature projectors into powerful LLMs, enabling them to understand and generate content based on visual and text input. Most of these models undergo two training steps, namely the pre-training stage and the fine-tuning stage. The main purpose of the pre-training stage is to align textual and visual features, while the goal of the fine-tuning stage is to bridge further the modality gap between vision and language, as well as lifting their instruction 
following capabilities and performance on specific downstream tasks. Early attempts, such as FLamingo \citep{Alayrac2022}, Gemini \citep{gemmateam2024gemma} ,and BLIP-2 \citep{Li2023b}, have already shown promising results. Recent works like LLaVA-v1.5 \citep{Liu2024}, Qwen2-VL \citep{Wang2024}, and xgen-mm \citep{Xue2024} have advanced this field, dramatically improving the abilities of these models. To enhance the consistency between vision representation and language representation, many efforts have been made, including utilizing higher resolution vision encoders, transitioning to larger and more capable LLMs, and adapting reinforcement learning techniques such as RLHF \citep{Ouyang2022, Yu2024}, among others. Despite the remarkable progress that has been made, LVLMs still suffer from severe hallucinations, which restricts their potential for large-scale applications in real-world scenarios.

\subsection{Mitigating Hallucinations in LVLMs}

Many efforts have been made to understand the causes of hallucination \citep{Liu2024b, Yin2023b, Bai2024, Lee2022DeduplicatingTD, Huang2023SurveyHL, Lin2022TruthfulQA, Venkit2023NationalityBI, Liu2023ExposingAGFFLM, Paullada2021DataDiscontentsAS, Li2023LargeLMCWM, Onoe2022EntityCBD, Liu2023ExposingAGFFLM, Aksitov2023CharacterizingAFTRA, Shi2023TrustingYEHLCD}. 
Existing approaches to mitigate hallucinations can be roughly divided into two categories based on the stage they take place. The first category focuses on the training phase. Most of the work falling into this category introduces additional or curated datasets \citep{Yu2024b, Liu2023, Gunasekar2023TextbooksAYN, Gunjal2024}, these works usually design a dataset specifically for hallucination-related tasks or improve data cleaning methods \citep{Raunak2021CuriousCH, Shen2021IdentifyingUS}, or introduce novel training objectives \citep{Lee2022FactualityEL, Shi2023InContextPLMBDB, chen2024alleviatinghallucinationslargevisionlanguage}. While effective, these methods typically require extensive training, which is both time-consuming and resource-intensive. The second category focuses on the inferencing phase, usually involves CD-based \citep{Li2023} novel decoding strategies \citep{Zhu2024MultilingualContrastiveDecoding, Chen2024HALC, Chuang2024DoLa, Zhong2024, Kim2024,woo2024dontmissforesttrees, kim2024codecontrastingselfgenerateddescription, phan2024distillationcontrastivedecodingimproving, zhao2024enhancingcontextualunderstandinglarge, Park2024, liu2024payingattentionimagetrainingfree}. Another approach is to detect potential hallucinations while generating and correct them \citep{Kuhn2023, Farquhar2024, Nikitin2024}, among others \citep{Yue2024LessIsM, Woo2024RITUAL, li2024inference, zhang2024truthxalleviatinghallucinationsediting}. In addition, some researchers tackle hallucinations by manipulating attention weight assigned to the image \citep{Zhu2024, Zhang2024, Wu2024NoiseBoost, huo2024selfintrospectivedecodingalleviatinghallucinations, xiao2024seeingimageprioritizingvisual}, or the prompt-related part of the image \citep{An2024}. Other works include prompt-based methods \citep{Qu2024LookCompareDecide, Wu2024LogicalClosedLoop, Xu2024ReReadingImprovesReasoning, Lee2024, Han2024, wang2024mitigatinghallucinationslargevisionlanguage}, utilizing external tools \citep{Chern2023FacTool, Yin2023, Zhao2024} or external knowledge \citep{Hu2024LRP4RAG, Qu2024AlleviatingHallucination, Ding2024AdaptiveRetrieval} and so on.\

However, one of the most vital features for large vision-language models, \ie the activation space in the inferencing stage, remains severely underexploited. Consequently, our study seeks to conduct head-level intervention during the inferencing stage, paving the way for more effective applications of large vision-language models.


\section{Task Formulation}
Given an LVLM parameterized by $ \theta $, the model processes a textual input $ \boldsymbol{X} = \{x_1, x_2, \dots, x_{N_x}\} $ and a visual input $ \boldsymbol{V} = \{v_1, v_2, \dots, v_{N_v}\} $, where $ N_x $ and $ N_v $ denote the sequence lengths of the text and visual inputs, respectively. The model then concatenates the textual and visual sequences to form a unified input $ \boldsymbol{H} = concat(\boldsymbol{Y},\boldsymbol{X}) $, which is subsequently passed through $L$ layers of a transformer architecture. At each layer, the concatenated input undergoes multi-head attention, which is computed as:
\begin{equation}
    \boldsymbol{H}^{(l+1)} = \boldsymbol{H}^{(l)} + \sum_{n=1}^{N} Attn_n^{(l)}(\boldsymbol{H}^{(l)}) \cdot W_o^{(l)},
\end{equation}
where $Attn_n^{(l)}$ denotes the attention operation of the $n$-th head at the $l$-th layer, and  $W_o^{(l)} \in \mathbb{R}^{ d \times Nd}$ is the output projection matrix, with $d$ representing the dimensionality of each attention head and $N$ the number of heads. Subsequently, the model autoregressively predicts the next token based on the output from the $L$-th layer: 
\begin{equation}
    p(y_t \mid y_{<t}) = Softmax(fc(\boldsymbol{H}_t^{(L)})),
\end{equation}
where $fc(\cdot)$ is an affine layer mapping the $Hd$-dimensional vector to a probability distribution of vocabulary size $M$.
\section{Methodology}
\begin{figure*}
  \centering
  \includegraphics[width=\textwidth]{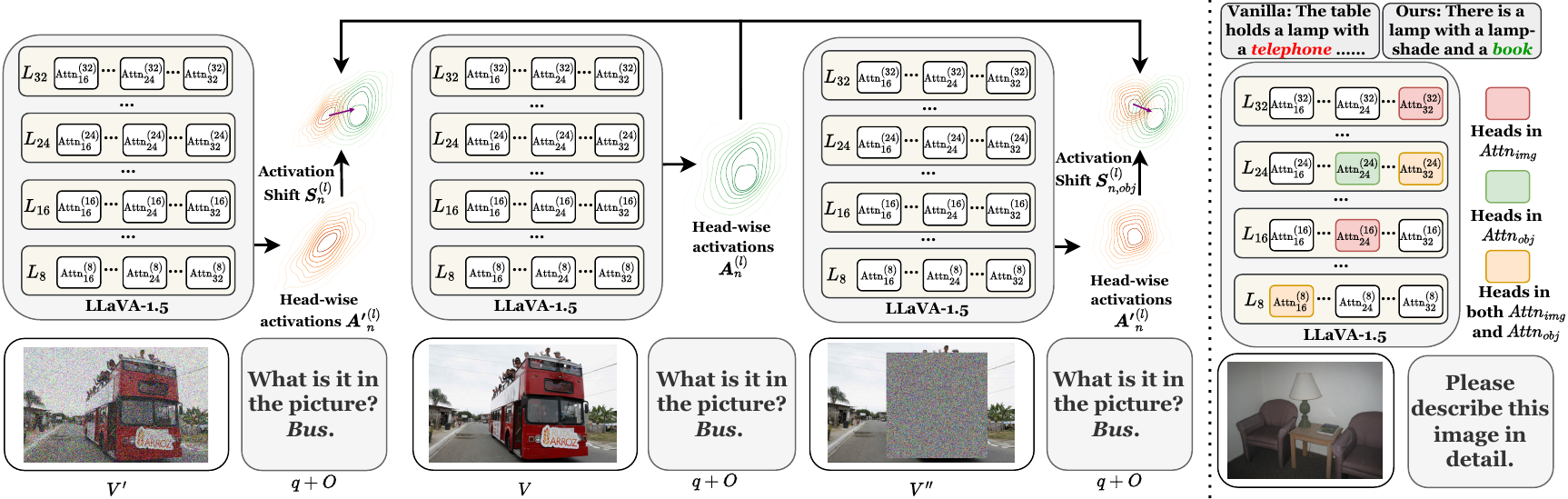}
  \vspace{-4mm}
  \caption{Overview of our proposed \modelname method. \modelname consists of two levels of intervention: Image-Level and Object-Level. The Image-Level module enhances the model’s focus on the overall scene to reduce reliance on language priors, while the Object-Level module focuses on fine-grained details to mitigate hallucinations. We apply targeted activation shifts to selected attention heads identified through binary classifiers trained to distinguish trusted and untrusted data pairs. }
  \label{figure:pipeline}
  \vspace{-5mm}
\end{figure*}
In this section, we introduce two levels of fine-grained intervention modules. The first module enables the LVLM to focus on the image, thereby reducing over-reliance on language priors. The second module encourages the LVLM to attend more closely to image objects, helping to mitigate the omission of critical objects and reduce hallucinations. 

\subsection{Intervention at Image-Level}
\label{sec:Image-Level}
This module aims to identify the attention heads associated with overall image information and to apply targeted interventions to these heads. This approach enhances the model's focus on visual input while diminishing the influence of language priors. 

As illustrated in Figure \ref{figure:pipeline}, consider a batch of image-question pairs, denoted as $\{(X_i, V_i)\}_{i=1}^{B}$, each labeled with the answer ``Yes'' from the POPE dataset. Here, $X_i$ takes the form of ``Is there a/an [object] in the image?'' For each question, we extract the specified object $O_i$ and reformulate the question to $q=$  ``What is it in the image?'' Then, for each image $V_i$, we progressively add Gaussian noise following the forward diffusion process \cite{ho2020denoising} to obtain the final blurred image $V_i'$:
\begin{equation}
\label{eq:noise}
\begin{aligned}
         q(V_i^{(t)} \mid V_i^{(t-1)}) &= \mathcal{N}(V_i^{(t)}; \sqrt{1 - \beta_t} V_i^{(t-1)}, \beta_t \mathbf{I}), \\
     q(V_i'\mid V_i) &= \prod_{t=1}^{T} q(V_i^{(t)} \mid V_i^{(t-1)}),
\end{aligned}
\end{equation}
where $\beta_t$ is the noise variance at step $t$. Finally, we construct a modified dataset $\{(q, V_i, V_i', O_i)\}_{i=1}^{B}$ to obtain Image-Level intervention vectors. We then concatenate $O_i$ with the question $q$ to form the answer, resulting in data pairs of trusted and untrusted data for the same sample: $(q + O_i, V_i)$ and $(q + O_i, V_i')$. For each sample, we treat the representation of the last token as the fused representation of multimodal data and extract the attention activation from a total of $N\times L$ heads across the $L$ layers for both trusted and untrusted data. These are denoted as $A_i = \{Attn_n^{(l)}\boldsymbol{H}^{(l)}_{s}\}_{n=1,l=1}^{N,L}$ and $A_i' = \{Attn_n^{(l)}\boldsymbol{H'}^{(l)}_{s}\}_{n=1,l=1}^{N,L}$, where $s=N_x+N_v+1$. According to the trusted activation $\{A_i\}_{i=1}^B$ and the untrusted activation $\{A_i'\}_{i=1}^B$ obtained from all sample pairs, we can calculate the following activation shift vector $\{\boldsymbol{S}_{n}^{(l)}\}_{n=1,l=1}^{N,L}$ that encourages the model to pay more attention to visual information, as follows:
\begin{equation}
   \boldsymbol{S}_{n}^{(l)} = \frac{1}{B} \sum_{i=1}^{B}\left(\boldsymbol{A}_{i,n}^{(l)} - \boldsymbol{A}_{i,n}'^{(l)}\right).
   \vspace{-2mm}
\end{equation}
Next, we train a binary classifier $f_{n}^{(l)}(\cdot)$ using $B$ sample pairs for each head to detect which heads encode Image-Level information, specifically those that can better distinguish the differences between pairs of trusted and untrusted samples. We then apply activation interventions to these selected heads:
\begin{equation}
    \begin{aligned}
        Attn_{\text{img}}\!&=\!\{Attn_n^{(l)}\!\mid\!Attn_n^{(l)}\in TopK(\text{Accuracy}(f_n^{(l)}(\cdot)))\},\\
        \boldsymbol{H}^{(l+1)}\! &=\! \boldsymbol{H}^{(l)}\!+\! \sum_{n=1}^{N} \left(Attn_n^{(l)}(\boldsymbol{H}^{(l)})\!+\! \mathbb{I}_{\text{img},n}^{(l)}\alpha\boldsymbol{S}_n^{(l)}\right)\!\cdot\!W_o^{(l)},
    \end{aligned}
\end{equation}
where $K$ is the number of selected intervention heads, $\mathbb{I}_{\text{img},n}^{(l)}$ is an indicator function, which is $1$ if $Attn_n^{(l)} \in Attn_{\text{img}}$ and $0$ otherwise, and $\alpha$ denotes the intensity of the intervention. 
After using $\{\boldsymbol{S}_{n}^{(l)}\}_{n=1,l=1}^{N,L}$ to perform Image-Level interventions on heads that encode image information, the model enhances the trustworthiness of the visual level, places greater attention on visual information, thus mitigates the impact of overly strong language priors.

\subsection{Intervention at Object-Level}
After enhancing the model's trustworthiness at the Image-Level, a more fine-grained, Object-Level intervention becomes necessary to increase the model's attention to image details, thereby reducing hallucinations caused by the omission of fine details. As shown in Figure \ref{figure:pipeline}, similar to Section \ref{sec:Image-Level}, we construct a dataset $\{(q, V_i, V_i'', O_i)\}_{i=1}^{B}$ to obtain Object-Level intervention vectors. We use Grounding DINO~\cite{liu2023grounding} to identify the area of object $O_i$ in image $V_i$. Gaussian noise is then added selectively to this object region, based on Eq. \ref{eq:noise}, producing a locally blurred image $V_i''$. Using both the original image $V_i$ and the partially blurred image $V_i''$, we construct trusted and untrusted data pairs for each sample: $(q + O_i, V_i)$ and $(q + O_i, V_i'')$ and analyze the attention activation values across $N \times L$ heads, allowing us to compute an Object-Level activation shift vector $\{\boldsymbol{S}_{n,\text{obj}}^{(l)}\}_{n=1,l=1}^{N,L}$:
\vspace{-3mm}
\begin{equation}
    \boldsymbol{S}_{n,\text{obj}}^{(l)} = \frac{1}{B} \sum_{i=1}^{B}\left(\boldsymbol{A}_{i,n}^{(l)} - \boldsymbol{A}_{i,n}''^{(l)}\right).
\end{equation}
\vspace{-2mm}

A binary classifier $f_{n,obj}^{(l)}(\cdot)$ is then trained to identify heads that effectively distinguish trusted from untrusted object-focused samples. Interventions are subsequently applied to the selected heads as follows:
\begin{equation}
    \begin{aligned}
        Attn_{\text{obj}}\!&=\!\{Attn_n^{(l)}\!\mid\!Attn_n^{(l)}\!\in\!TopK(\text{Accuracy}(f_{n,obj}^{(l)}(\cdot)))\},\\
        \boldsymbol{H}^{(l+1)}\!&=\!\boldsymbol{H}^{(l)}\!+\! \sum_{n=1}^{N}\!\left(Attn_n^{(l)}(\boldsymbol{H}^{(l)})\!+\! \mathbb{I}_{\text{obj},n}^{(l)} \beta \boldsymbol{S}_{\text{obj},n}^{(l)}\right)\!\cdot\!W_o^{(l)},
    \end{aligned}
\end{equation}
where $K$ is the number of intervention heads, $\mathbb{I}_{\text{obj},n}^{(l)}$ is an indicator function set to $1$ if $Attn_n^{(l)} \in Attn_{\text{obj}}$ and $0$ otherwise, and $\beta$ controls the intervention intensity.

Finally, we integrate the Image-Level and Object-Level intervention modules to create a unified approach that reinforces the model’s focus on both the overall visual context and finer object-specific details as follows:
\begin{equation}
    \begin{aligned}
        \boldsymbol{H}^{(l+1)} &= \boldsymbol{H}^{(l)}+ \sum_{n=1}^{N} \Big( Attn_n^{(l)} (\boldsymbol{H}^{(l)}) \\ 
        &+ \mathbb{I}_{\text{img},n}^{(l)} \alpha \boldsymbol{S}_{n}^{(l)} 
        + \mathbb{I}_{\text{obj},n}^{(l)} \beta \boldsymbol{S}_{\text{obj},n}^{(l)} \Big) \cdot W_o^{(l)}.
    \end{aligned}
\end{equation}

By applying these interventions in tandem, the model gains a balanced attention mechanism that mitigates excessive reliance on language priors while enhancing sensitivity to essential visual cues across different levels of granularity, thereby alleviating the occurrence of hallucinations.

\section{Experiments}
\subsection{Experimental Setup}
\noindent \textbf{Datasets and Metrics.}

\noindent\textbf{POPE} \cite{li2023evaluating} (Polling-based Object Probing Evaluation) is a benchmark designed to assess how effectively LVLMs recognize the presence of specific objects in images, thus identifying Object-Level hallucinations. It uses Yes/No questions based on object annotations, with metrics including Accuracy, Precision, Recall, and F1 score. The dataset is balanced, with 50\% of queries for existing objects and 50\% for non-existing ones, and employs three sampling strategies: random, popular, and adversarial. Drawing from MSCOCO~\cite{lin2014microsoft}, A-OKVQA \cite{schwenk2022okvqa}, and GQA~\cite{hudson2019gqa} datasets, POPE evaluates 27,000 query-answer pairs to gauge model performance. 

\noindent \textbf{MME} \cite{fu2023mme} (Multimodal Large Language Model Evaluation) benchmark is designed to comprehensively evaluate the performance of LVLMs across various dimensions. It includes ten perception-focused tasks and four cognition-related tasks. MME benchmark specifically addresses Object-Level hallucinations through subsets focusing on object existence and count, 
while attribute-level hallucinations are assessed via subsets related to object position and color. The evaluation metric is Accuracy, providing a quantitative measure of the model's performance across these diverse tasks.

\noindent \textbf{Baselines.} We adopt the widely used LLaVA-v1.5 \cite{liu2023improvedllava} and Qwen-VL~\cite{Bai2023} models as our baseline LVLMs.
We compared two baselines that eliminate language priors of the LVLMs in the decoding stage to alleviate hallucinations: VCD \cite{leng2024mitigating} and Opera \cite{huang2024opera}.

\noindent \textbf{Implementation Details.} In our experiments, we utilized 1,500 QA pairs with ``Yes'' responses from the COCO Random subset of the POPE dataset to train the intervention shift vector. Subsequently, we evaluated the \modelname approach on two datasets with significant distributional differences: POPE, and MME. This evaluation was conducted to assess the generalizability and robustness of \modelname across diverse data distributions. For each attention head, we employed Support Vector Machines (SVMs)~\cite{cortes1995support} as the classifier and performed 2-fold cross-validation to evaluate the classification accuracy. In the experiment, we set $\alpha = \beta$ and determined the optimal values of $\alpha$, $\beta$, and $K$ through a grid search. Detailed hyperparameter configurations are provided in Appendix A. All experiments were conducted on a system equipped with $8 \times \text{H800}$ GPUs.

\subsection{Main Results}
\paragraph{Results on POPE.} Table \ref{tab:pope-results} presents the results of LLaVA-v1.5 and Qwen-VL on nine subsets of the POPE dataset. From comparing these methods, we can derive the following conclusions: \textbf{1)} Applying \modelname results in an average
 
\noindent  performance improvement on the F1 score for LLaVA-v1.5  and  Qwen-VL of 7.09\% and 5.44\% on 9 subsets respectively, which are respectively higher than the previous contrastive decoding sota baseline (Opera) 2.19\% and 1.14\%. This improvement can be attributed to the fact that \modelname does not eliminate language priors, which may provide useful information. Instead, it enhances the model’s attention to various levels of visual information, thereby reducing the model’s tendency to overly rely on language priors and mitigating the occurrence of hallucinations. \textbf{2)} Individually intervening at both the image level and object level, with average gains of 5.76\% and 5.47\% in F1 score, respectively. This demonstrates that enhancing LVLMs' attention to image information at various levels can effectively mitigate hallucinations. Moreover, since Object-Level intervention also implicitly directs the model’s focus toward broader image information, it achieves a comparatively higher performance gain. \textbf{3)} The intervention shift vector trained using 1,500 samples from the MSCOCO random subset achieves an average improvement on F1 score of 7.67\% on the MSCOCO random subset and an average improvement of 6.09\% across the other eight subsets with different distributions. This indicates that \modelname generalizes effectively, and the resulting intervention vector captures a general direction towards trustworthiness rather than merely fitting a specific dataset.
\vspace{-4mm}

\paragraph{Results on MME.}

\begin{figure}[h!]
  \centering
  \includegraphics[width=\linewidth]{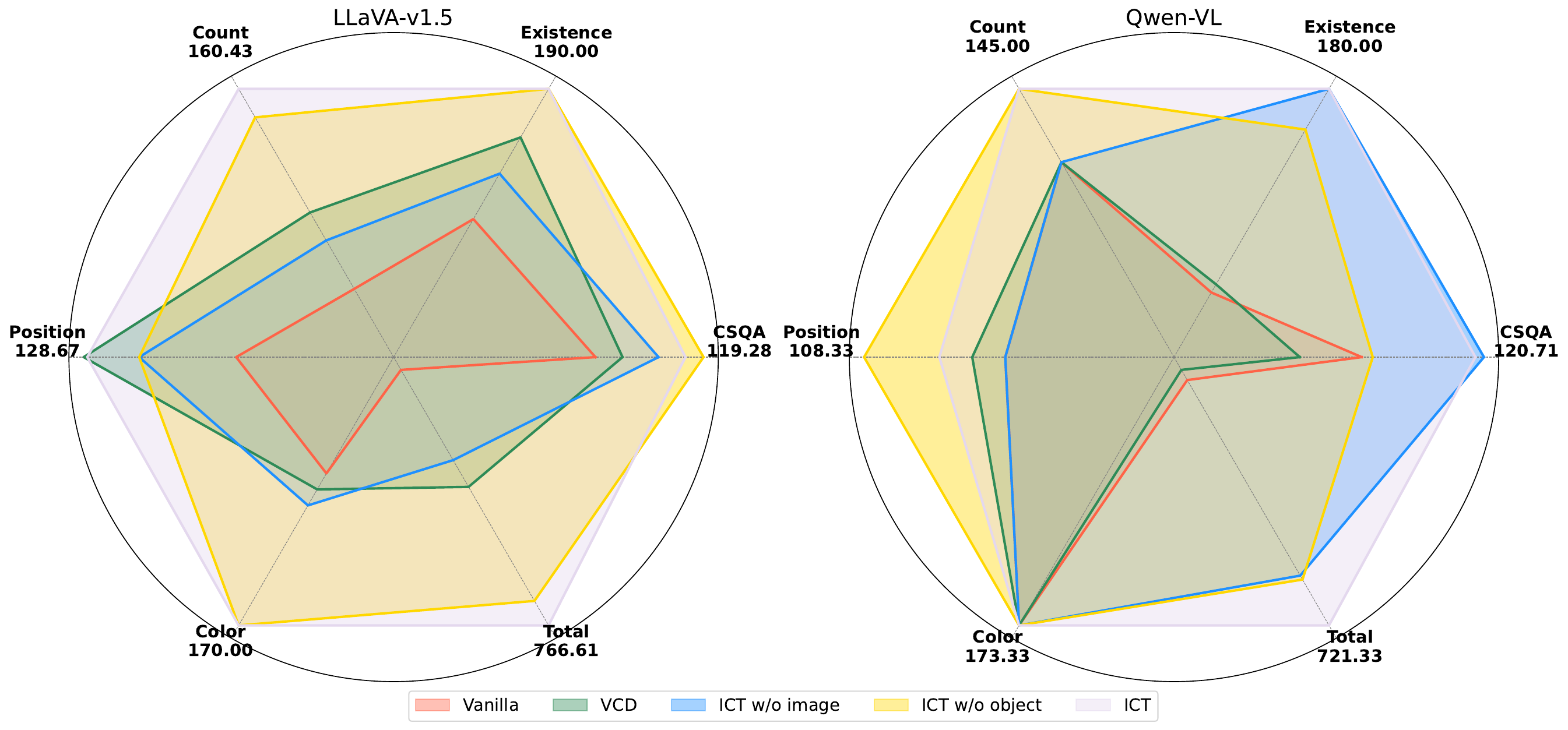}
  \vspace{-6mm}
\caption{Comparison of \modelname with baseline methods (Vanilla and VCD) on the MME benchmark. The radar chart illustrates improvements across various evaluation categories, including existence, position, count, color, and commonsense QA (CSQA). }
  \label{fig:mme}
  \vspace{-6mm}
\end{figure}

Figure \ref{fig:mme} presents the experimental results on the MME benchmark, where we followed \citet{leng2024mitigating} evaluation setup, focusing on the hallucination subset of MME. Compared to the unmodified model, \modelname achieved a score improvement of 80.51 on LLaVA-v1.5 and 34.67 on Qwen-VL, surpassing the performance of VCD. Additionally, on non-hallucination tasks such as commonsense QA, \modelname demonstrated a 10-point improvement on LLaVA-v1.5 and a 9.5-point improvement on Qwen-VL. These results indicate that \modelname not only effectively reduces hallucinations but also enhances general reasoning capabilities across different models. Moreover, since the objects present in the images of the MME dataset are generally prominent, the variant \modelname~\textit{w/o} image (i.e., without the Image-Level intervention that enhances the model’s attention to the overall scene) provides only limited improvements compared to \modelname \textit{w/o} object. This demonstrates that the absence of Image-Level guidance restricts the model's ability to fully leverage global visual context, thereby reducing its effectiveness in scenarios where the dominant objects play a crucial role.

\begin{table*}[]
\centering

\begin{adjustbox}{width=0.85\linewidth}
\tiny
    \begin{tabular}{ccccccc}
    \toprule
        \textbf{Dataset}  & \textbf{Setting}  & \textbf{Method}  & \multicolumn{2}{c}{\textbf{LLaVA-v1.5}}  &  \multicolumn{2}{c}{\textbf{Qwen-VL}}  \\ 
\cmidrule(lr){4-5}
\cmidrule(lr){6-7}
        ~ & ~ & ~ & Accuracy & F1 Score & Accuracy & F1 Score \\ 
        \midrule
        \multirow{18}{*}{COCO}  & \multirow{6}{*}{\textit{Random}}  & Regular & 83.29  & 81.33  & 84.37 & 82.67 \\ 
        ~ & ~ & VCD & 87.73  & 87.16  & 88.63 & 87.81 \\ 
        ~ & ~ & OPERA & 89.20  & 88.81  &  87.31&  86.92\\ 
        ~ & ~ & \modelname \textit{w/o} image & \underline{89.70 \textcolor{OliveGreen}{(+6.41)} } & \underline{89.93 \textcolor{OliveGreen}{(+8.60)}}  & 88.76\textcolor{OliveGreen}{(+4.39)} & 87.84\textcolor{OliveGreen}{(+5.17)} \\ 
        ~ & ~ & \modelname \textit{w/o} object & 88.50 \textcolor{OliveGreen}{(+5.21)}  & 88.81 \textcolor{OliveGreen}{(+7.48)}  & \underline{89.13\textcolor{OliveGreen}{(+4.76)}} & \underline{88.20\textcolor{OliveGreen}{(+5.63)}} \\ 
        ~ & ~ & \modelname & \textbf{90.11 \textcolor{OliveGreen}{(+6.82)}}  & \textbf{90.03 \textcolor{OliveGreen}{(+8.70)}}  & \textbf{89.46\textcolor{OliveGreen}{(+5.09)}} & \textbf{89.20\textcolor{OliveGreen}{(+6.63)}} \\ 
        \cline{2-7}
        ~ & \multirow{6}{*}{\textit{Popular}}  & Regular & 81.88  & 80.06  & 84.13 & 82.06 \\ 
        ~ & ~ & VCD & 85.38  & 85.06  & 87.12 & 86.40 \\ 
        ~ & ~ & OPERA & 86.64  & 86.62  & 87.44& 86.68\\ 
        ~ & ~ & \modelname \textit{w/o} image & \underline{87.43 \textcolor{OliveGreen}{(+5.55)}}  & \underline{87.09 \textcolor{OliveGreen}{(+7.03)}}  & \underline{87.63\textcolor{OliveGreen}{(+3.50)}} & \underline{86.78\textcolor{OliveGreen}{(+4.72)}} \\ 
        ~ & ~ & \modelname \textit{w/o} object & 87.30 \textcolor{OliveGreen}{(+5.42)}  & 86.65 \textcolor{OliveGreen}{(+6.59)}  & 87.10\textcolor{OliveGreen}{(+2.97)} & 86.28\textcolor{OliveGreen}{(+4.22)} \\ 
        ~ & ~ & \modelname & \textbf{87.50 \textcolor{OliveGreen}{(+5.62)} } & \textbf{87.60 \textcolor{OliveGreen}{(+7.54)}}  & \textbf{88.16\textcolor{OliveGreen}{(+4.03)}} & \textbf{87.33\textcolor{OliveGreen}{(+5.27)}} \\ 
        \cline{2-7}
        ~ & \multirow{6}{*}{\textit{Adversarial}}  & Regular & 78.96  & 77.57  & 82.26 & 80.37 \\ 
        ~ & ~ & VCD & 80.88  & 81.33  & 84.26 & 83.90 \\ 
        ~ & ~ & OPERA & 81.24  & 81.38  &  84.78&  83.45\\ 
        ~ & ~ & \modelname \textit{w/o} image & \underline{84.13 \textcolor{OliveGreen}{(+5.17)}}  &  \underline{83.31 \textcolor{OliveGreen}{(+5.74)}}  & \underline{85.13\textcolor{OliveGreen}{(+2.87)}} & 84.35\textcolor{OliveGreen}{(+3.98)} \\ 
        ~ & ~ & \modelname \textit{w/o} object & 83.63 \textcolor{OliveGreen}{(+4.67)}  & 82.13 \textcolor{OliveGreen}{(+4.56)}  & \textbf{85.23\textcolor{OliveGreen}{(+2.97)}} & \textbf{84.62\textcolor{OliveGreen}{(+4.25)}} \\ 
        ~ & ~ & \modelname & \textbf{84.43 \textcolor{OliveGreen}{(+5.47)} } & \textbf{83.74 \textcolor{OliveGreen}{(+6.17)} } & 84.96\textcolor{OliveGreen}{(+2.70)} & \underline{84.42\textcolor{OliveGreen}{(+4.05)}} \\ 
        \midrule
        \multirow{18}{*}{AOKVQA}  & \multirow{6}{*}{\textit{Random}}  & Regular & 83.45  & 82.56  & 86.67 & 85.59 \\ 
        ~ & ~ & VCD & 86.15  & 86.34  & 89.22 & 89.01 \\ 
        ~ & ~ & OPERA & 88.02  & 84.59  & 88.19& 88.43\\ 
        ~ & ~ & \modelname \textit{w/o} image & \textbf{89.60 \textcolor{OliveGreen}{(+6.15)}}  & \textbf{89.75 \textcolor{OliveGreen}{(+7.19)} } & \textbf{89.46\textcolor{OliveGreen}{(+2.79)}} & \textbf{89.28\textcolor{OliveGreen}{(+3.69)}} \\ 
        ~ & ~ & \modelname \textit{w/o} object & 89.00 \textcolor{OliveGreen}{(+5.55)}  & 88.91 \textcolor{OliveGreen}{(+6.35)}  & 89.26\textcolor{OliveGreen}{(+2.59)} & 88.95\textcolor{OliveGreen}{(+3.36)} \\ 
        ~ & ~ & \modelname &  \raisebox{-2ex}{\textbf{}}\underline{89.20 \textcolor{OliveGreen}{(+5.75)}}  &  \underline{89.41 \textcolor{OliveGreen}{(+6.85)}}  & \textbf{89.46\textcolor{OliveGreen}{(+2.79)}} & \underline{89.03\textcolor{OliveGreen}{(+3.44)}} \\ 
        \cline{2-7}
        
        ~ & \multirow{6}{*}{\textit{Popular}}  & Regular & 79.90  & 79.59  & 85.56 & 84.63 \\ 
        ~ & ~ & VCD & 81.85  & 82.82  & 87.85 & \underline{87.81} \\ 
        ~ & ~ & OPERA & 83.22  &  \underline{84.67}  & 87.91& 87.13\\ 
        ~ & ~ & \modelname \textit{w/o} image &  \underline{85.36 \textcolor{OliveGreen}{(+5.46)} } &\textbf{ 85.34 \textcolor{OliveGreen}{(+5.75)} } & 87.70\textcolor{OliveGreen}{(+2.14)} & 87.68\textcolor{OliveGreen}{(+3.05)} \\ 
        ~ & ~ & \modelname \textit{w/o} object & 84.65 \textcolor{OliveGreen}{(+4.75)}  & 84.32 \textcolor{OliveGreen}{(+4.73)}  & \underline{87.90\textcolor{OliveGreen}{(+2.34)}} & 87.48\textcolor{OliveGreen}{(+2.85)} \\ 
        ~ & ~ & \modelname &\textbf{ 85.73 \textcolor{OliveGreen}{(+5.83)}}  & \textbf{85.34 \textcolor{OliveGreen}{(+5.75)}  }& \textbf{88.13\textcolor{OliveGreen}{(+3.57)}} & \textbf{87.83\textcolor{OliveGreen}{(+3.20)}} \\ 
        \cline{2-7}
        
        ~ & \multirow{6}{*}{\textit{Adversarial}}  & Regular & 74.04  & 75.15  & 79.57 & 79.50 \\ 
        ~ & ~ & VCD & 74.97  & 77.73  & 81.27 & \underline{82.38} \\ 
        ~ & ~ & OPERA & 73.82  & 77.91  & 80.82& 81.54\\ 
        ~ & ~ & \modelname \textit{w/o} image & 79.26 \textcolor{OliveGreen}{(+5.22)}& \underline{80.14 \textcolor{OliveGreen}{(+4.99)}}  & \underline{81.50\textcolor{OliveGreen}{(+1.93)} }& 82.10\textcolor{OliveGreen}{(+2.60)} \\ 
        
        ~ & ~ & \modelname \textit{w/o} object & 77.79 \textcolor{OliveGreen}{(+3.75)}  & 79.53 \textcolor{OliveGreen}{(+4.38)}  & 81.40\textcolor{OliveGreen}{(+1.83)} & 82.18\textcolor{OliveGreen}{(+2.68)} \\ 
        
        ~ & ~ & \modelname & \textbf{79.60 \textcolor{OliveGreen}{(+5.56)} } &  \textbf{80.43 \textcolor{OliveGreen}{(+5.28)}}  & \textbf{81.94\textcolor{OliveGreen}{(+2.37)}}& \textbf{82.44\textcolor{OliveGreen}{(+2.94)}}\\ 
        \midrule
        
        \multirow{18}{*}{GQA}  & \multirow{6}{*}{\textit{Random}}  & Regular & 83.73  & 82.95  & 80.97 & 79.01 \\ 
        ~ & ~ & VCD & 86.65  & 86.99  & 85.59 & 85.33 \\ 
        ~ & ~ & OPERA & 88.13  & 88.91  & 86.02& 85.29\\ 
        ~ & ~ & \modelname \textit{\textit{w/o}} image &  \underline{89.03 \textcolor{OliveGreen}{(+5.30)} } & 88.99 \textcolor{OliveGreen}{(+6.04)}  & 85.20\textcolor{OliveGreen}{(+4.23)} & 85.90\textcolor{OliveGreen}{(+6.89)} \\ 
        ~ & ~ & \modelname \textit{w/o} object & 88.97 \textcolor{OliveGreen}{(+5.24)}  &  \underline{89.30 \textcolor{OliveGreen}{(+6.35)}}  & \textbf{87.20\textcolor{OliveGreen}{(+6.23)}} & \underline{86.80\textcolor{OliveGreen}{(+7.79)}} \\ 
        ~ & ~ & \modelname & \textbf{89.60 \textcolor{OliveGreen}{(+5.87)}  }&\textbf{ 89.44 \textcolor{OliveGreen}{(+6.49)}  }& \raisebox{-2ex}{\textbf{}}\underline{86.38\textcolor{OliveGreen}{(+5.41)}} & \textbf{86.96\textcolor{OliveGreen}{(+7.95)}} \\ 
        \cline{2-7}
        
        ~ & \multirow{6}{*}{\textit{Popular}}  &  Regular & 78.17  & 78.37  & 75.99 & 74.84 \\ 
        ~ & ~ & VCD & 80.73  & 82.24  & 81.83 & \underline{82.23} \\ 
        ~ & ~ & OPERA & 79.27  & 82.11  & 81.97& 82.12\\ 
        ~ & ~ & \modelname \textit{w/o} image &  \underline{84.23 \textcolor{OliveGreen}{(+6.06)}}  & 83.92 \textcolor{OliveGreen}{(+5.55)}  & 81.43\textcolor{OliveGreen}{(+5.44)} & 80.86\textcolor{OliveGreen}{(+6.02)} \\ 
        ~ & ~ & \modelname \textit{w/o} object &\textbf{ 84.70 \textcolor{OliveGreen}{(+6.53)} } & \underline{ 84.45 \textcolor{OliveGreen}{(+6.08)} } & \textbf{83.93\textcolor{OliveGreen}{(+7.94)}} & \textbf{82.95\textcolor{OliveGreen}{(+8.11)}} \\ 
        ~ & ~ & \modelname & \textbf{84.70 \textcolor{OliveGreen}{(+6.53)}  }&\textbf{ 84.78 \textcolor{OliveGreen}{(+6.41)}}  & 
        \raisebox{-2ex}{\textbf{}}\underline{82.63\textcolor{OliveGreen}{(+6.64)}} & 82.22\textcolor{OliveGreen}{(+7.38)} \\ 
       \cline{2-7}
        
        ~ & \multirow{6}{*}{\textit{Adversarial}}  & Regular & 75.08  & 76.06  & 75.46 & 74.33 \\ 
        ~ & ~ & VCD & 76.09  & 78.78  & 80.01 & \underline{80.75} \\ 
        ~ & ~ & OPERA & 75.00  & 78.71  & 80.24& 80.64\\ 
        ~ & ~ & \modelname \textit{w/o} image & \textbf{81.92 \textcolor{OliveGreen}{(+6.84)}  }& \underline{82.11 \textcolor{OliveGreen}{(+6.05)}}  & 80.03\textcolor{OliveGreen}{(+4.57)} & 79.90\textcolor{OliveGreen}{(+5.47)} \\ 
        ~ & ~ & \modelname \textit{w/o} object &  \underline{81.65 \textcolor{OliveGreen}{(+6.57)}}  & 82.00 \textcolor{OliveGreen}{(+5.94)}  & \textbf{81.60\textcolor{OliveGreen}{(+6.14)}} & \textbf{81.53\textcolor{OliveGreen}{(+7.10)}} \\ 
        ~ & ~ & \modelname & 81.50 \textcolor{OliveGreen}{(+6.42)}  & \textbf{82.27 \textcolor{OliveGreen}{(+6.21)}}  & \underline{80.83\textcolor{OliveGreen}{(+5.37)}} & 80.60\textcolor{OliveGreen}{(+6.17)} \\ 
        \bottomrule
    \end{tabular}

\end{adjustbox}
\caption{\textbf{Main results on POPE tasks.} We evaluate the accuracy of various LVLMs on the POPE task across the MSCOCO, A-OKVQA, and GQA datasets. \textbf{Regular} represents the setting where direct sampling is applied. \textbf{ICT \textit{w/o} image} and \textbf{ICT \textit{w/o} object} correspond to the exclusion of image-level and object-level interventions, respectively. The \textbf{bold} and \underline{underlined} values indicate the highest and second-highest metrics under each setting, respectively.}
\label{tab:pope-results}
\vspace{5mm}
\end{table*}
\newpage

\section{Analysis and Discussions}
\subsection{Inference Latency across Different Methods}

In Table \ref{tab:efficiency}, we compare the inference efficiency of two different representative approaches with our proposed method. Notably, while the CD-based methods result in a significant increase in inference time (2.1 to 3.6 times slower), our \modelname incurs virtually no impact on inference efficiency. We attribute this to the fact that both VCD and OPERA either require multiple inference runs or involve substantial additional computations during inference. In contrast, \modelname introduces no extra computational cost at inference, further highlighting the advantages of our method.
\begin{table}[!htbp] 
\begin{center}
\small
\vspace{-1em}
\resizebox{0.98 \linewidth}{!}{
\begin{tabular}{lrrr}
\toprule
\bf Method & \bf20-Token Len & \bf50-Token Len & \bf80-Token Len \\ \midrule
LLaVA-v1.5  & 405.3 \color{gray}$\uparrow$$\times$1.0 &934.6  \color{gray}$\uparrow$$\times$1.0 & 1440.0 \color{gray}$\uparrow$$\times$1.0 \\ 
 + VCD   & 988.9 {\color{Maroon}$\uparrow$} \color{Maroon}$\times$ {\color{Maroon}2.4} & 2031.7 {\color{Maroon}$\uparrow$} \color{Maroon}$\times$ {\color{Maroon}2.2}  & 3077.5 {\color{Maroon}$\uparrow$} \color{Maroon}$\times$ {\color{Maroon}2.1}  \\
 + OPERA & 1371.7 {\color{Maroon}$\uparrow$} \color{Maroon}$\times$ {\color{Maroon}3.4} & 3294.1 \color{Maroon}{$\uparrow$} $\times$ {3.5}  & 5717.3 {\color{Maroon}$\uparrow$} \color{Maroon}$\times$ {\color{Maroon}3.6}  \\ \midrule 
\bf+ \modelname  & \bf415.9 {\color{OliveGreen}$\uparrow$} \color{OliveGreen}$\times$ {\color{OliveGreen}1.0} & \bf931.9 {\color{OliveGreen}$\uparrow$} \color{OliveGreen}$\times$ {\color{OliveGreen}1.0}  & \bf1485.5 {\color{OliveGreen}$\uparrow$} \color{OliveGreen}$\times$ {\color{OliveGreen}1.0}   \\ \bottomrule

\end{tabular}
}
\vspace{-1.em}
\caption{\small Comparison of the efficiency of different methods in generating tokens of varying lengths on an NVIDIA H800 GPU. Inference times are recorded in milliseconds.}
\label{tab:efficiency}
\end{center}
\vspace{-2.6em}
\end{table}
\FloatBarrier

\subsection{Generalizability between Different Models}
\begin{figure}
  \centering
  \includegraphics[width=\linewidth]{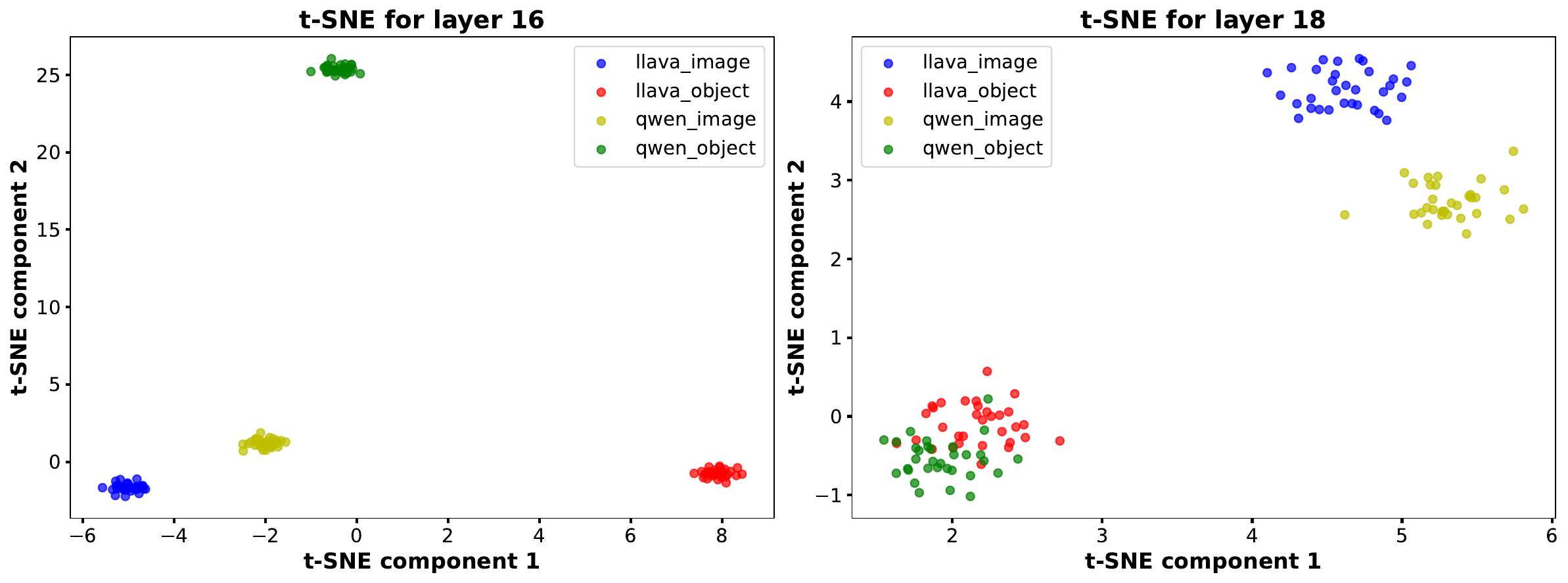}
  \vspace{-6mm}
    \caption{t-SNE visualization of Object-Level and Image-Level offset vectors for LLaVA-v1.5 and Qwen-VL at layers 16 and 18.}
  \label{figure:tsne}
  \vspace{-8mm}
\end{figure}

To verify that the derived activation shift vectors to some extent represent the model's alignment with authenticity, we explored whether these shifts are transferable to other models. Specifically, we applied the activation shift vectors obtained from LLaVA-v1.5 on the COCO random subset to intervene in Qwen-VL. We then evaluated the generalizability of this intervention on the GQA dataset, which has a distribution significantly different from COCO. The results in Table \ref{tab:cross_model} demonstrate that \modelname-LLaVA-v1.5 achieves an average improvement of $4.62\%$ in F1 score and $4.64\%$ in accuracy compared to the unmodified model. Additionally, we employed t-SNE to visualize the offset vectors across different layers for LLaVA-v1.5 and Qwen-VL. Figure \ref{figure:tsne} shows the Object-Level and Image-Level shift vectors of multi-head attention in the 16th and 18th layers for both models. From the figure, we can observe that in layer 16, the Image-Level vectors (blue and yellow) are relatively close to each other, indicating a degree of similarity in encoding Image-Level information between the models. However, the Object-Level vectors (red and green) are more distinct, reflecting model-specific characteristics in encoding fine-grained object details. In layer 18, the Image-Level and Object-Level vectors for both models exhibit high similarity, indicating that the resulting activation shift vectors represent universal shift vectors that guide the model to pay more attention to visual information.
\vspace{-4mm}
\begin{table}[b]
\centering
\vspace{-5mm}

\begin{adjustbox}{width=0.85\linewidth}
    \begin{tabular}{cccc}
    \toprule
         \textbf{Setting}  & \textbf{Method}   & Accuracy & F1 Score \\ 
        \midrule
         \multirow{3}{*}{\textit{Random}}  & Regular & 80.97  & 79.01  \\ 
         ~ & VCD & 85.59  & 85.33 \\ 
         ~ & \modelname & 86.96  & 86.38 \\ 
         ~ & \modelname-LLaVA-v1.5 & 85.10  & 83.27 \\ 
         \midrule
         
         \multirow{3}{*}{\textit{Popular}}  & Regular & 75.99  & 74.84 \\
         ~ & VCD &  81.83  & 82.23  \\ 
         ~ & \modelname & 82.63  & 82.22 \\ 
         ~ & \modelname-LLaVA-v1.5 & 81.50  & 80.10 \\ 
         \midrule
         
          \multirow{3}{*}{\textit{Adversarial}}  & Regular & 75.46  & 74.33  \\ 
         ~ & VCD & 80.01  & 80.75 \\ 
         ~ & \modelname & 80.83  & 80.60 \\ 
         ~ & \modelname-LLaVA-v1.5 & 79.73  & 78.68 \\ 
    \bottomrule
         
    \end{tabular}

\end{adjustbox}
\caption{\textbf{Cross-model generalization of \modelname.} Performance of Qwen-VL on GQA after applying activation shift from LLaVA-v1.5. \modelname-LLaVA-v1.5 refers to the results obtained by applying LLaVA-v1.5’s activation shift vectors to intervene in Qwen-VL.}

\label{tab:cross_model}
\vspace{-5mm}
\end{table}

\begin{figure*}[t]
  \centering
  \includegraphics[width=0.83\linewidth]{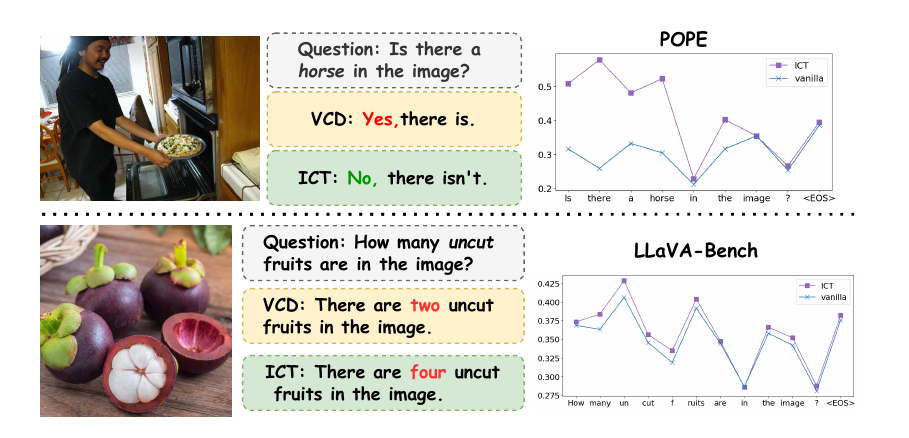}
  \vspace{-6mm}
    \caption{Case Study and Error Analysis of \modelname.}
  \label{figure:casestudy}
  \vspace{-6mm}
\end{figure*}
\subsection{Impact of Hyperparameters on Performance}
\begin{figure}[htbp]
  \centering
  \includegraphics[width=\linewidth]{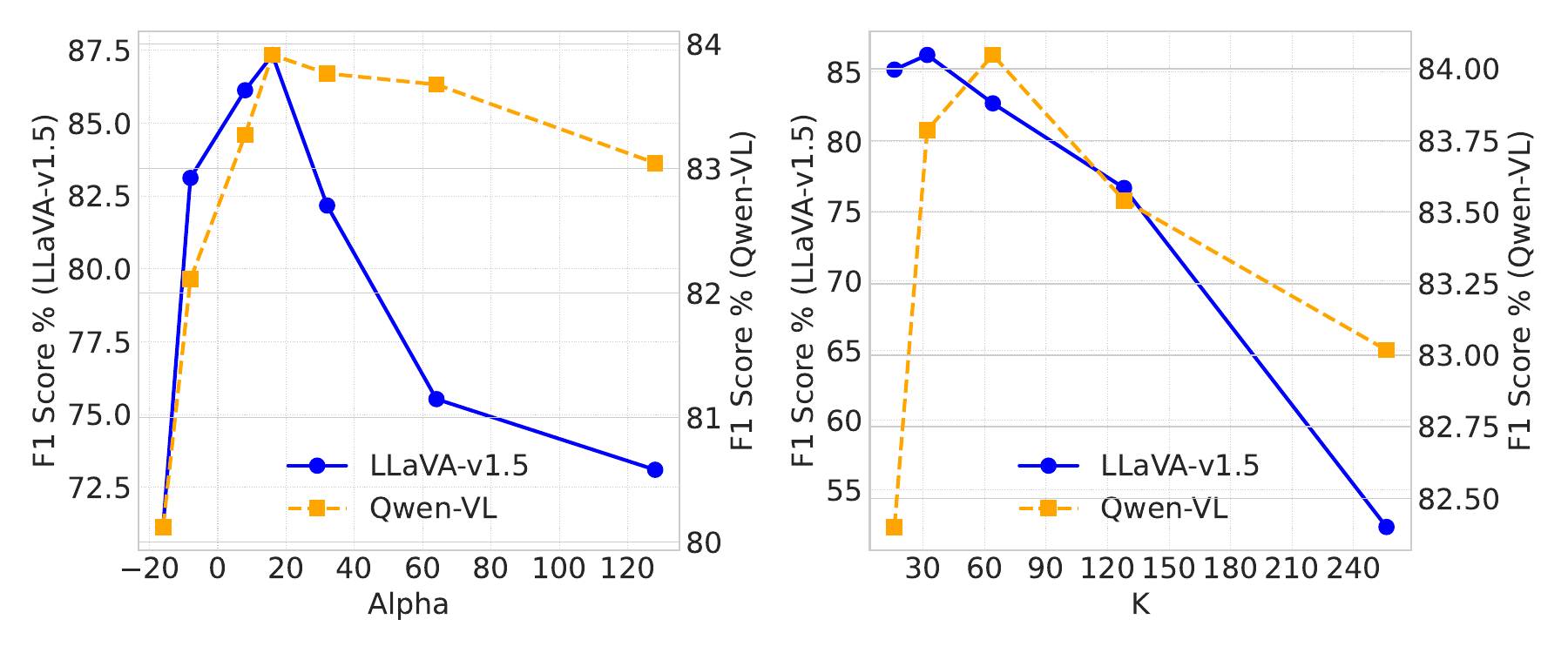}
  \vspace{-6mm}
    \caption{Effect of hyperparameters $\alpha$ (left) and $K$ (right) on the F1 score for LLaVA-v1.5 and Qwen-VL models on the POPE COCO random subset.}
  \label{figure:para}
  \vspace{-8mm}
\end{figure}

\begin{figure}[b]
  \centering
  \includegraphics[width=\linewidth]{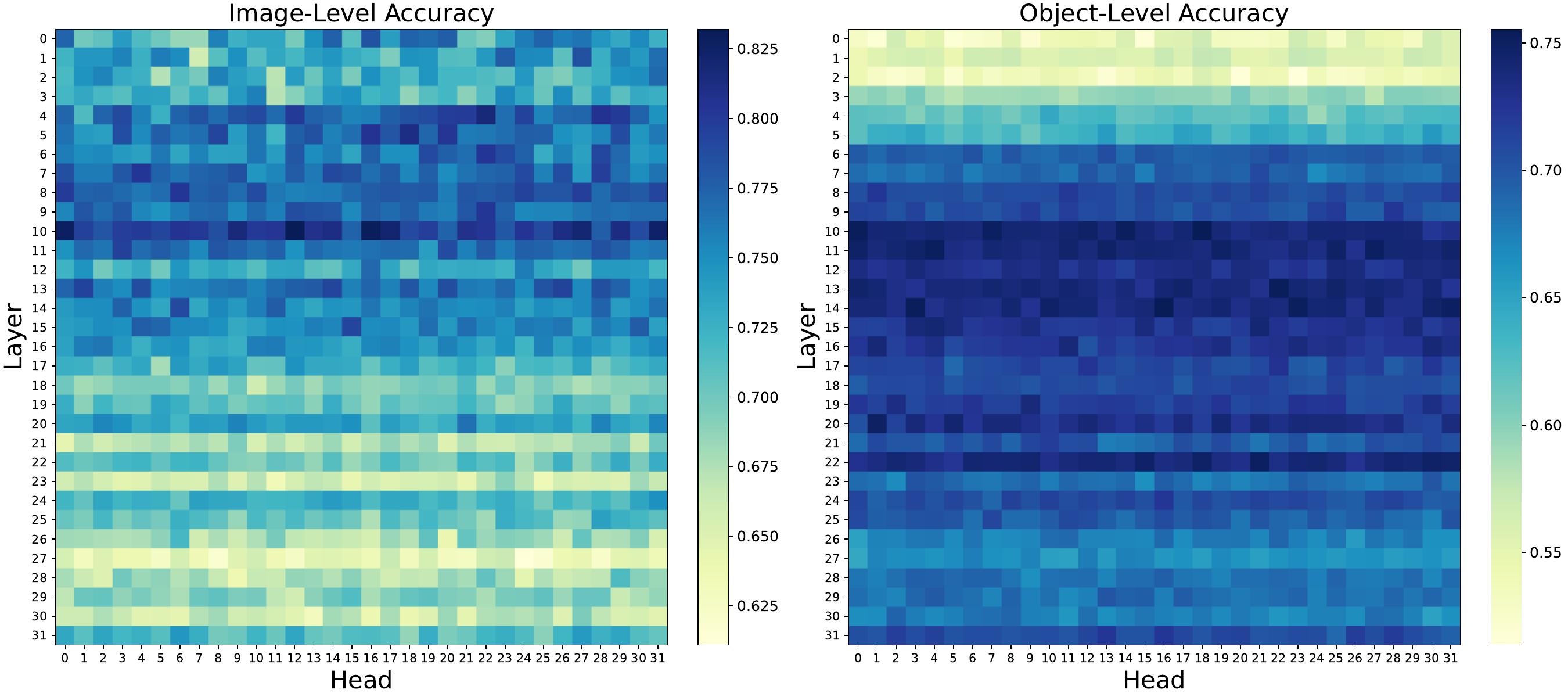}
  \vspace{-6mm}
  \caption{Heatmap of classification accuracy for Image-Level (left) and Object-Level (right) features of LLaVA-v1.5.}
  \label{figure:heatmap}
  \vspace{-8mm}
\end{figure}
Our proposed \modelname primarily relies on two key hyperparameters: the intervention intensity $\alpha$ and the number of heads $K$ involved in the intervention. To investigate the impact of these hyperparameters on performance, we fixed one of the parameters and calculated the average F1 score for LLaVA-v1.5 and Qwen-VL on the POPE COCO random subset. The results are presented in Figure \ref{figure:para}, we can observe that when $\alpha$ is negative—indicating that reverse intervention reduces the model's attention to visual information—the model's performance declines. When $\alpha$ is too small, the model does not receive sufficient intervention, resulting in suboptimal outcomes. Conversely, when $\alpha$ is excessively large, the intervention becomes too strong, disrupting the model’s foundational capabilities and leading to a decrease in performance. For the hyperparameter $K$, we find that when $K$ is too small, certain attention heads that encode relevant visual information are not adequately intervened upon, resulting in suboptimal performance. Conversely, when $K$ is too large, attention heads that encode irrelevant information are unnecessarily interfered with, which leads to performance degradation.
\subsection{Analysis of Attention Heads for Visual Information Encoding}
In Figure \ref{figure:heatmap}, we present the classification accuracy of the binary classifiers for Image-Level and Object-Level features across the 1,024 multi-head attention heads in the 32 layers of LLaVA-v1.5. This analysis helps identify which heads encode overall visual information and which heads capture fine-grained visual details. From the figure, we can observe that the attention heads encoding overall visual information are predominantly located in the earlier layers of the model, such as layers 4 and 10. In contrast, the attention heads that capture fine-grained visual details are more concentrated in the later layers, such as layers 20 and 22.

\subsection{Case Study and Error Analysis}
In Figure \ref{figure:casestudy}, we present a case study on the POPE and LLaVA-Bench datasets, visualizing the attention ratio of each question text token to the visual tokens. As illustrated, after applying \modelname, the model allocates a higher proportion of attention to visual tokens, especially to object tokens relevant to the question (e.g., ``horse'' and ``fruits''). By prioritizing visual information, \modelname correctly identifies the absence of a horse in the image, whereas VCD erroneously concludes that a horse is present due to insufficient attention to visual cues. However, when asked, "How many uncut fruits are in the picture?", VCD incorrectly answered ``two'' due to a lack of focus on visual details. Although \modelname correctly identifies that there are a total of four fruits in the image, this question requires not only attention to visual content but also reasoning within the text modality. The model needs to not only recognize the overall count of fruits but also focus on the attribute ``uncut''. Due to its failure to incorporate this information, \modelname outputs a wrong answer.

\vspace{-2mm}
\section{Conclusion and Limitations}
In this paper, we mitigate the issue of object hallucinations in LVLMs by proposing Image-Object Cross-Level Trusted Intervention (\modelname), a training-free, plug-and-play method that enhances the model’s focus on both Image-Level and Object-Level visual information during the forward pass. Our experiments show that \modelname significantly reduces hallucinations while also improving general reasoning capabilities. The results confirm that \modelname effectively mitigates excessive reliance on language priors, enhancing both accuracy and robustness in diverse visual contexts. 

\noindent \textbf{Limitations.} Our method requires access to the model’s weights, so it cannot be applied to closed-source models. Additionally, we only use Gaussian blur as the transformation method on the image. Future research could explore the use of generative methods for transforming images.

{
    \small
    \bibliographystyle{ieeenat_fullname}
    \bibliography{main}
}
\clearpage
\setcounter{page}{1}

\maketitlesupplementary

\section*{Appendix A: Implementation Details
}
\begin{figure*}[t]
  \centering
  \includegraphics[width=\textwidth]{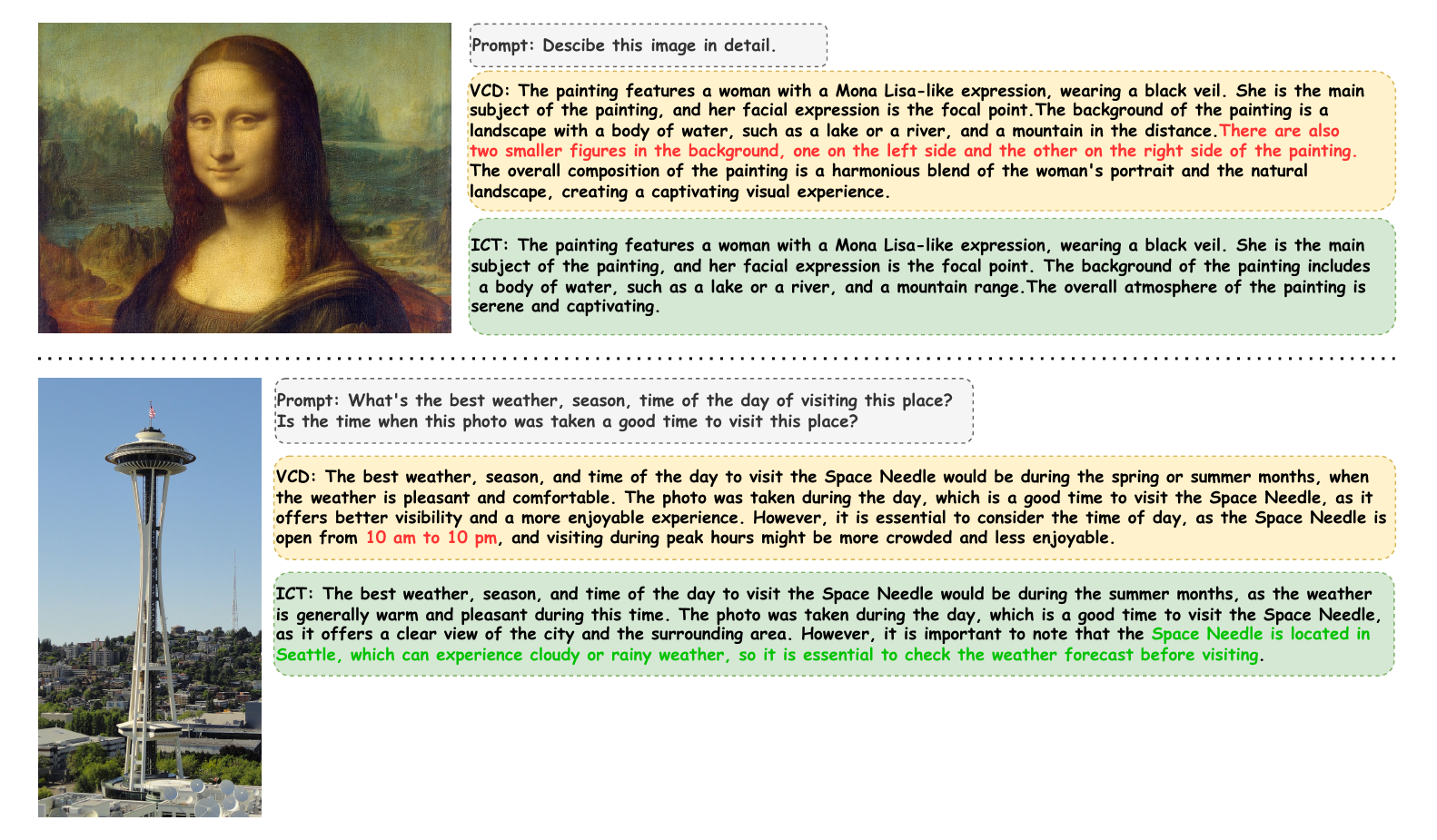}
  \vspace{-4mm}
  \caption{More case studies of \modelname. In the first example, our \modelname avoids captioning non-existent objects in the image while keeping the rest of the caption nearly identical. This demonstrates its ability to refine the accuracy of generated content while maintaining the quality and coherence of the text. In the second example, VCD hallucinated the opening hours of the Space Needle, whereas our \modelname correctly retrieved this information from its internal knowledge, stored implicitly within the language modality. This highlights our method's remarkable capability to minimize vision-related hallucinations while leveraging the advantages of language priors.}
  \label{figure:case-app}
  \vspace{-5mm}
\end{figure*}

In our experiments, we used greedy search to ensure reproducibility. Hyperparameter tuning was performed using a grid search to explore the possible combinations of key parameters systematically. The hyperparameters under consideration included $\alpha$, $\beta$, and $K$. To reduce the dimensionality of the search space, we constrained $\alpha$ and $\beta$ to be equal. This constraint is reasonable as these two hyperparameters serve similar roles in our experiments. For LLaVA-v1.5, we applied interventions to the layers \texttt{language\_model.model.layers.\{i\}.self\_attn .o\_proj}, while for Qwen-VL, the interventions were applied to \texttt{transformer.h.\{i\}.attn.c\_proj}.

\paragraph{Grid Search Process}
The grid search exhaustively evaluated all combinations of hyperparameters within the specified ranges. Given the aforementioned constraint on $\alpha$ and $\beta$, the search space was defined as the Cartesian product:
\[
\{(8, 8), (16, 16), (24, 24), (32, 32)\} \times \{32, 64, 128, 256\}.
\]
This resulted in a total of $4 \times 4 = 16$ unique hyperparameter configurations, which is a relatively small search space.

\subsubsection*{Hyperparameter Search Details}

\paragraph{$\alpha$ and $\beta$:}
The hyperparameters $\alpha$ and $\beta$, which control the strength of Image-Level and Object-Level intervention, respectively, were allowed to take values from the discrete set $\{8, 16, 24, 32\}$. This range was chosen to strike a balance between improving model trustworthiness and maintaining overall performance. By enforcing the constraint $\alpha = \beta$, the number of unique combinations was reduced to four, thereby simplifying the search space while ensuring sufficient exploration of the parameter landscape.

\paragraph{$K$:}
The hyperparameter $K$, representing the number of attention heads we intervene on, was explored over the set $\{32, 64, 128, 256\}$. This range was determined based on two key considerations:
\begin{itemize}
    \item Both LLaVA-v1.5 and Qwen-VL architectures consist of 32 layers, each with 32 attention heads, resulting in 1024 total attention heads. The chosen range allows for intervention on varying subsets of these attention heads.
    \item Preliminary classification accuracy results, as shown in Figure \ref{figure:heatmap}, indicate that this range provides sufficient flexibility to cover scenarios from intervening on all heads that are relevant to hallucination to intervening only on those most pertinent to truthfulness.
\end{itemize}

\section*{Appendix B: More Case Studies}
In Figure \ref{figure:case-app}, we present additional illustrative cases from LLaVA-bench to further demonstrate the effectiveness of \modelname on extremely challenging open-ended questions. The upper part of Figure \ref{figure:case-app} vividly illustrates the precision and reliability of our intervention. \modelname retains most of the original phrasing, selectively eliminating segments containing untruthful content. This demonstrates that \modelname not only significantly reduces the model's tendency to hallucinate but also achieves this with minimal and minimal side effects.

The lower part of Figure \ref{figure:case-app} presents another example where \modelname successfully preserves and utilizes useful language priors, whereas VCD fails to do so. In this case, both approaches accurately identified the building, but VCD hallucinated incorrect scene opening hours. In contrast, \modelname provided suggestions grounded in factual knowledge, leveraging the positive aspects of language priors effectively.

\end{document}